\documentclass[11pt,a4paper]{article}

\usepackage[T1]{fontenc}
\usepackage[utf8]{inputenc}
\usepackage{lmodern}
\usepackage[margin=2.5cm]{geometry}
\usepackage{amsmath,amssymb}
\usepackage{graphicx}
\usepackage{xcolor}
\usepackage{hyperref}
\usepackage{booktabs}
\usepackage{array}
\usepackage{colortbl}
\usepackage{microtype}
\usepackage{setspace}
\usepackage{natbib}
\usepackage{tikz}
\usetikzlibrary{
  positioning, arrows.meta, shapes.geometric, shapes.misc,
  fit, backgrounds, calc, decorations.pathreplacing, matrix
}
\usepackage{pgfplots}
\pgfplotsset{compat=1.18}

\definecolor{llmblue}{HTML}{3949AB}
\definecolor{mtpblue}{HTML}{5C6BC0}
\definecolor{latentblue}{HTML}{7986CB}
\definecolor{wmbg}{HTML}{E8EAF6}
\definecolor{lpbg}{HTML}{C5CAE9}
\definecolor{mtpbg}{HTML}{9FA8DA}
\definecolor{ntpbg}{HTML}{7986CB}
\definecolor{specA}{HTML}{E3F2FD}
\definecolor{specB}{HTML}{BBDEFB}
\definecolor{specC}{HTML}{90CAF9}
\definecolor{specD}{HTML}{42A5F5}
\definecolor{specE}{HTML}{1565C0}
\definecolor{nodegrn}{HTML}{2E7D32}
\definecolor{nodeblu}{HTML}{1565C0}

\hypersetup{
  colorlinks=true, linkcolor=llmblue, citecolor=llmblue, urlcolor=llmblue
}

\title{\textbf{From Tokens to States: LLMs as a Special Case of\\World Models and the Continuous Path Beyond}}
\author{Paul Dubois}
\date{Opinion Paper, Draft, June 2026}

\begin{document}
\maketitle
\thispagestyle{empty}

\begin{abstract}
The AI community has framed the relationship between large language models (LLMs) and world models as a dichotomy: LLMs predict tokens; world models simulate reality.
\citet{lecun2022path} argues that reaching general intelligence requires abandoning autoregressive token prediction in favour of latent-space architectures.
This framing is unnecessarily binary.
Two claims will be defended.
First, LLMs are a degenerate special case of world models: the state space is the set of all token sequences, the only action is appending one token, and world models are therefore a strict generalisation of LLMs, not a replacement.
Second, there is a natural continuous spectrum from NTP to JEPA, with multi-token prediction, future-summary prediction, and next-latent prediction as intermediate stations already populated by current research.
Moving along this spectrum relaxes the LLM constraints one by one.
It also progressively surrenders the two practical advantages that make LLMs trainable at scale:
internet-scale self-supervised data, and a transformer architecture co-designed for discrete token prediction.
Both are examined as open research questions:
the \emph{data question} (the cliff from self-supervised text to instrumented action-labelled environments) and the \emph{architecture question}
(whether the transformer generalises to continuous-state prediction, or whether a new primitive is needed).
\end{abstract}

\newpage
\tableofcontents
\newpage

\section{Introduction}

A \emph{world model} tracks the state of an environment and predicts how it evolves over time.
It supports planning by simulating the consequences of actions before committing to them \citep{ha2018worldmodels, lecun2022path}.
LLMs, trained to predict the next token in a sequence, look like an entirely different object.
\citeauthor{lecun2022path}'s (\citeyear{lecun2022path}) position paper formalises this intuition.
His Joint Embedding Predictive Architecture (JEPA) predicts in latent space rather than token space, conditioned on agent actions.
He explicitly contrasts ``word models'' with ``world models'' and argues that token prediction cannot yield the structured representations needed for planning and reasoning.

This framing is imprecise.
``World model'' is not the name of a specific architecture; it is a formal class parameterised by a state space, an action space, and a transition function.
LLMs are a member of this class with specific, restrictive choices for each parameter.
The implication is not that LLMs are good enough, but that the path to more powerful world models is a progressive relaxation of those choices, not a clean break.

The key empirical support for this view comes from mechanistic interpretability: LLMs trained only on token sequences demonstrably build internal world-model representations in their hidden activations.
\citet{li2023othello} and \citet{nanda2023othello} show that OthelloGPT encodes the full board state linearly in its activations at $>$99\% accuracy, despite seeing only move tokens.
Chess-playing LMs replicate this \citep{karvonen2024chess}.
Llama-2 encodes linear representations of geographic space and calendar time \citep{gurnee2024space}.
These results show that the world model is in the activations, not the tokens; the tokens are the interface, not the representation.

The continuous-spectrum view is also supported by recent architectures:
multi-token prediction \citep{gloeckle2024mtp}, future-summary prediction \citep{mahajan2025future}, and next-latent prediction \citep{teoh2025nextlat} each populate a point between standard NTP and JEPA.
\citet{li2026word} empirically confirm that LLMs can function as text-based world models at sufficient scale.
No prior work makes the formal containment claim or unifies these architectures under a single spectrum argument.

\section{LLMs as a Special Case of World Models}

\subsection{World Model Formalism}

\begin{figure}[h]
\centering
\resizebox{\textwidth}{!}{%
\begin{tikzpicture}[
  node distance=1.6cm and 2.0cm,
  state/.style={rectangle, rounded corners=4pt, draw=llmblue!60,
                fill=llmblue!10, minimum width=2.0cm, minimum height=0.8cm,
                align=center, font=\small},
  trans/.style={rectangle, rounded corners=4pt,
                draw={rgb,255:red,74;green,144;blue,217},
                fill={rgb,255:red,74;green,144;blue,217},
                text=white, minimum width=2.6cm, minimum height=0.8cm,
                align=center, font=\small},
  arr/.style={-{Stealth[length=5pt]}, thick, gray!60},
  lbl/.style={font=\footnotesize\itshape, text=gray!80, midway, above=2pt}
]
  \node[state] (s0) {State $s_t$};
  \node[trans, right=of s0] (T0) {Transition\\$T(s_t,a_t)$};
  \node[state, right=of T0] (s1) {State $s_{t+1}$};
  \node[trans, right=of s1] (T1) {Transition\\$T(s_{t+1},a_{t+1})$};
  \node[state, right=of T1] (s2) {State $s_{t+2}$};

  \draw[arr] (s0) -- node[lbl]{action $a_t$} (T0);
  \draw[arr] (T0) -- node[lbl]{sample} (s1);
  \draw[arr] (s1) -- node[lbl]{action $a_{t+1}$} (T1);
  \draw[arr] (T1) -- node[lbl]{sample} (s2);
\end{tikzpicture}}
\caption{A world model iteratively applies the transition function to simulate multi-step trajectories.
State and action spaces are left intentionally abstract.}
\label{fig:wm}
\end{figure}
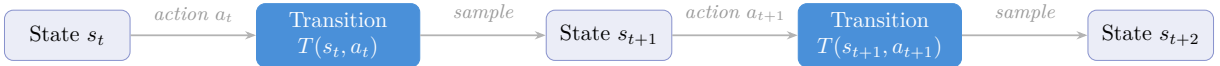

A world model is a tuple $(S, A, T, \rho_0)$: $S$ is the state space, $A$ the action space, $T : S \times A \to \mathcal{P}(S)$ the transition function, and $\rho_0$ the initial state distribution.
Planning proceeds by iterative querying of $T$ to simulate a trajectory $(s_0, a_0, s_1, a_1, \ldots)$ in imagination before committing to real actions \citep{ha2018worldmodels} (Figure~\ref{fig:wm}).

\subsection{Formal Embedding of LLMs}

\begin{quote}
\textbf{Claim:} Every autoregressive LLM can be formally embedded in the world-model formalism, making world models a strict generalisation of LLMs: $\text{LLMs} \subset \text{World Models}$.
\end{quote}

\begin{figure}[h]
\centering
\begin{tikzpicture}[
  box/.style={rectangle, rounded corners=4pt, draw=llmblue!60,
              fill=llmblue!10, text width=3.2cm, align=center,
              minimum height=1.1cm, font=\small},
  act/.style={rectangle, rounded corners=4pt, draw=orange!70,
              fill=orange!15, text width=2.4cm, align=center,
              minimum height=1.1cm, font=\small},
  arr/.style={-{Stealth[length=5pt]}, thick},
  lbl/.style={font=\footnotesize\itshape, midway, above=2pt}
]
  \node[box] (s0) {State $s_t$\\[2pt]\footnotesize$=$ token sequence\\[2pt]\footnotesize\texttt{"The cat sat on the"}};
  \node[act, right=3.2cm of s0] (a) {Action $a_t$\\[2pt]\footnotesize$=$ next token\\[2pt]\footnotesize\texttt{"mat"}};
  \node[box, right=3.2cm of a] (s1) {State $s_{t+1}$\\[2pt]\footnotesize \texttt{"The cat sat on the mat"}};

  \draw[arr, llmblue!70] (s0) -- node[lbl]{LLM policy $\pi(\cdot|s_t)$} (a);
  \draw[arr, orange!70]  (a)  -- node[lbl]{deterministic append} (s1);
\end{tikzpicture}
\caption{The LLM as world model: state $=$ token sequence, action $=$ next token, transition $=$ deterministic append.}
\label{fig:llm-wm}
\end{figure}
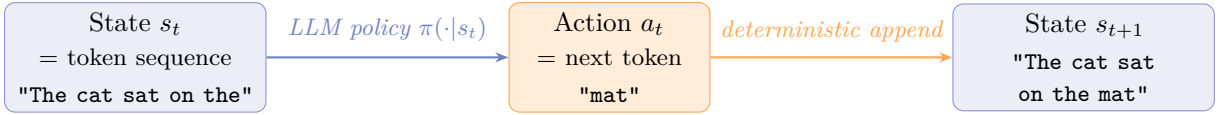

Let $V$ be the vocabulary (a finite set of tokens).
Define:
\begin{itemize}
  \item $S = V^*$: the state space is the set of all finite token sequences (the current context)
  \item $A = V$: the only ``action'' is choosing the next token
  \item $T(s, a) = \delta_{s \cdot a}$: the transition is \emph{deterministic}, appending token $a$ to sequence $s$ yields $s \cdot a$
  \item The LLM provides $\pi_\theta : V^* \to \mathcal{P}(V)$: the policy for selecting the next action
\end{itemize}
Figure~\ref{fig:llm-wm} illustrates this mapping concretely.

One subtlety: in a standard world model, $T$ encodes external world dynamics and $\pi$ is the agent.
In the LLM case, the transition is trivial (append) and all content is in the policy.
The LLM simultaneously acts as world simulator and agent.
This conflation is a defining feature of the LLM special case.

\subsection{LLM Constraints Within the World-Model Class}

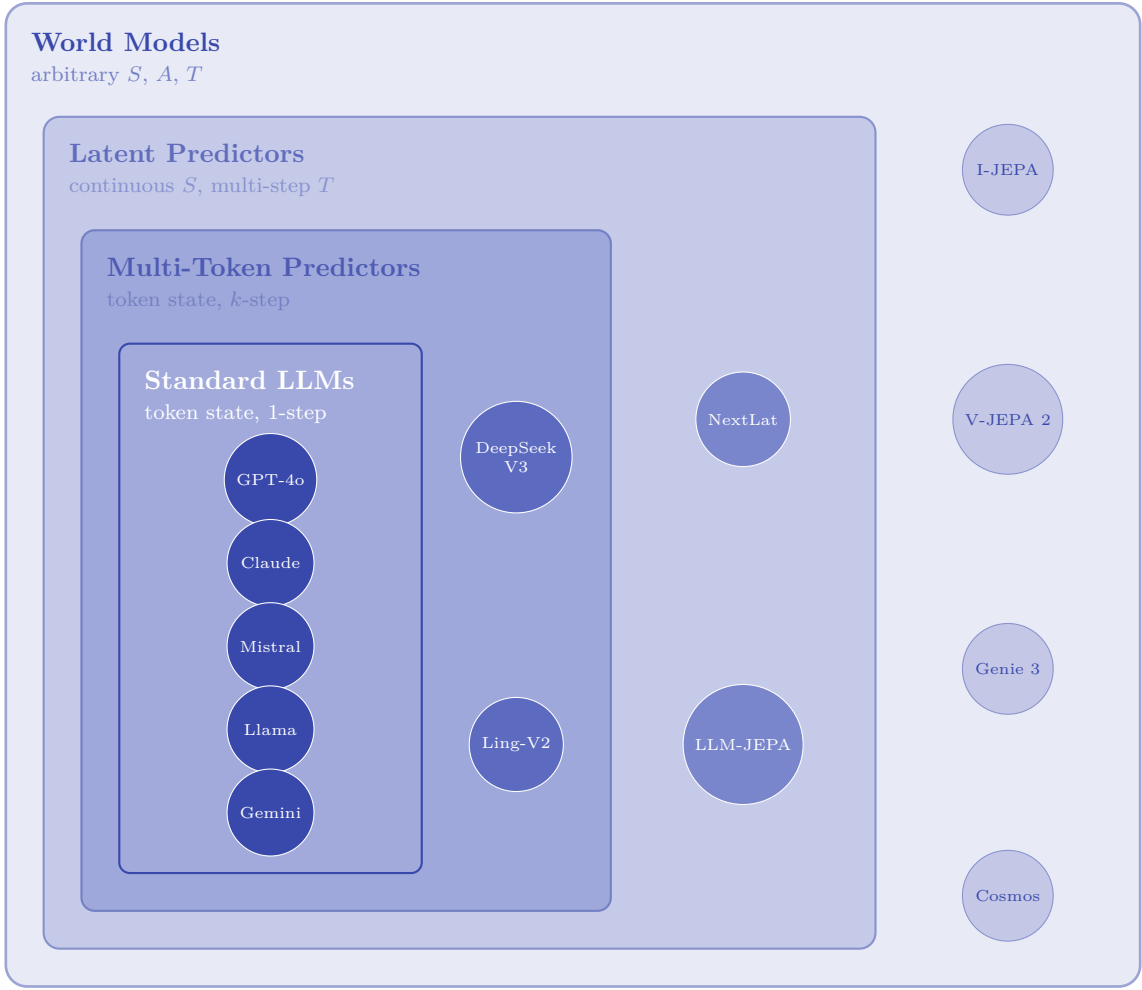
\begin{figure}[ht]
\centering
\begin{tikzpicture}[x=1cm, y=1cm]

\fill[wmbg, draw=llmblue!50, line width=1pt, rounded corners=8pt]
  (0,0) rectangle (15,13);
\node[anchor=north west, font=\small\bfseries, text=llmblue]
  at (0.2,12.75) {World Models};
\node[anchor=north west, font=\scriptsize, text=llmblue!70]
  at (0.2,12.3) {arbitrary $S$, $A$, $T$};

\fill[lpbg, draw=llmblue!60, line width=0.8pt, rounded corners=6pt]
  (0.5,0.5) rectangle (11.5,11.5);
\node[anchor=north west, font=\small\bfseries, text=llmblue!80]
  at (0.7,11.28) {Latent Predictors};
\node[anchor=north west, font=\scriptsize, text=llmblue!60]
  at (0.7,10.83) {continuous $S$, multi-step $T$};

\fill[mtpbg, draw=llmblue!70, line width=0.8pt, rounded corners=5pt]
  (1.0,1.0) rectangle (8.0,10.0);
\node[anchor=north west, font=\small\bfseries, text=llmblue!90]
  at (1.2,9.78) {Multi-Token Predictors};
\node[anchor=north west, font=\scriptsize, text=llmblue!70]
  at (1.2,9.33) {token state, $k$-step};

\fill[ntpbg!70, draw=llmblue, line width=0.8pt, rounded corners=4pt]
  (1.5,1.5) rectangle (5.5,8.5);
\node[anchor=north west, font=\small\bfseries, text=white]
  at (1.7,8.28) {Standard LLMs};
\node[anchor=north west, font=\scriptsize, text=white!85]
  at (1.7,7.83) {token state, 1-step};

\foreach \lbl/\cy in {GPT-4o/6.7, Claude/5.6, Mistral/4.5, Llama/3.4, Gemini/2.3}{
  \node[circle, fill=llmblue, draw=white, text=white,
        minimum size=1.15cm, font=\tiny, align=center] at (3.5,\cy) {\lbl};
}

\node[circle, fill=mtpblue, draw=white, text=white,
      minimum size=1.2cm, font=\tiny, align=center]
  at (6.75, 7.0) {DeepSeek\\V3};
\node[circle, fill=mtpblue, draw=white, text=white,
      minimum size=1.2cm, font=\tiny, align=center]
  at (6.75, 3.2) {Ling-V2};

\node[circle, fill=latentblue, draw=white, text=white,
      minimum size=1.2cm, font=\tiny, align=center]
  at (9.75, 7.5) {NextLat};
\node[circle, fill=latentblue, draw=white, text=white,
      minimum size=1.2cm, font=\tiny, align=center]
  at (9.75, 3.2) {LLM-JEPA};

\node[circle, fill=llmblue!30, draw=llmblue!60, text=llmblue,
      minimum size=1.2cm, font=\tiny, align=center]
  at (13.25, 10.8) {I-JEPA};
\node[circle, fill=llmblue!30, draw=llmblue!60, text=llmblue,
      minimum size=1.2cm, font=\tiny, align=center]
  at (13.25, 7.5) {V-JEPA 2};
\node[circle, fill=llmblue!30, draw=llmblue!60, text=llmblue,
      minimum size=1.2cm, font=\tiny, align=center]
  at (13.25, 4.2) {Genie 3};
\node[circle, fill=llmblue!30, draw=llmblue!60, text=llmblue,
      minimum size=1.2cm, font=\tiny, align=center]
  at (13.25, 1.2) {Cosmos};

\end{tikzpicture}
\caption{Containment hierarchy as nested sets.
LLMs (innermost box) are the most constrained special case.
Each outer ring relaxes one constraint.
This is a \emph{subset} relation, not a replacement.}
\label{fig:containment}
\end{figure}

The containment is strict because general world models over continuous state spaces, conditioned on external actions, are not LLMs.
Figure~\ref{fig:containment} shows the resulting subset hierarchy.
Compared to a general world model, LLMs impose five constraints (Table~\ref{tab:constraints}):

\newcommand{\rowA}{\rowcolor{gray!12}}
\newcommand{\rowB}{\rowcolor{white}}
\begin{table}[h]
\centering
\renewcommand{\arraystretch}{1.4}
\begin{tabular}{@{} p{3.4cm} p{5.2cm} p{5.0cm} @{}}
\toprule
\textbf{Constraint} & \textbf{General World Model} & \textbf{LLM} \\
\midrule
\rowA State representation
  & Arbitrary (continuous, multimodal, latent)
  & Discrete token sequences only \\
\rowB Action granularity
  & Semantic agent actions
  & One micro-token per step \\
\rowA State persistence
  & Explicit stored variable
  & Re-derived from context each pass \\
\rowB Training supervision
  & Paired (state, action, next state) tuples
  & Next-token prediction on raw text \\
\rowA Layer architecture
  & Open (SSMs, diffusion nets, hierarchical encoders)
  & Transformer, co-designed for discrete tokens \\
\bottomrule
\end{tabular}
\caption{The five constraints that characterise LLMs as a special case of world models.}
\label{tab:constraints}
\end{table}

The last two constraints are tightly coupled and explain why LLMs have scaled so well:
NTP on text requires no labels, no sensors, no instrumented environments; only internet text ($\sim$10$^{13}$ tokens).
The transformer architecture was co-evolved with this objective, making the pairing exceptionally efficient.
These are not bugs, but they are constraints, and relaxing them is what moving up the spectrum means.

\subsection{Empirical Support: World Models Inside LLMs}
\label{sec:empirical}

\begin{figure}[h]
\centering
\begin{tikzpicture}[
  layer/.style={rectangle, rounded corners=4pt, minimum width=9cm,
                minimum height=1.3cm, align=center, font=\small},
  arr/.style={-{Stealth[length=5pt]}, thick, gray!60},
  lbl/.style={font=\footnotesize\itshape, text=gray}
]
  \node[layer, fill=nodeblu, text=white] (emb) at (0,0)
    {\textbf{Input embedding layer}\\
     Move tokens $\to$ token vectors};
  \node[layer, fill=nodegrn!80, text=white, below=0.5cm of emb] (trf)
    {\textbf{Transformer layers}\\[2pt]
     \footnotesize Hidden activations encode world state:\\
     board position $\cdot$ piece ownership $\cdot$ legal moves\\
     \textit{(linearly decodable at $>$99\% accuracy)}};
  \node[layer, fill=nodeblu, text=white, below=0.5cm of trf] (head)
    {\textbf{Output head}\\
     Hidden state $\to$ next move distribution};

  \draw[arr] (emb) -- node[lbl, right]{residual stream} (trf);
  \draw[arr] (trf) -- node[lbl, right]{residual stream} (head);

  \node[font=\footnotesize, text=nodegrn!70, align=right,
        left=1.2cm of trf] (wmlbl) {world model\\lives here};
  \draw[-{Stealth[length=7pt]}, thick, nodegrn!70] (wmlbl.east) -- (trf.west);
  \draw[decorate, decoration={brace, amplitude=5pt}, thick, gray!50]
    ([xshift=4.9cm]emb.north) -- ([xshift=4.9cm]head.south)
    node[midway, right=6pt, font=\footnotesize, text=gray!70,
         align=left] {token\\interface};
\end{tikzpicture}
\caption{OthelloGPT is a standard transformer trained only on move tokens.
The token interface (blue) handles discrete move symbols; the transformer layers (green) are where the world model resides.
This maps directly to Claim~1:
tokens are the action space $A = V$; the latent world state lives in the hidden activations.}
\label{fig:othello}
\end{figure}
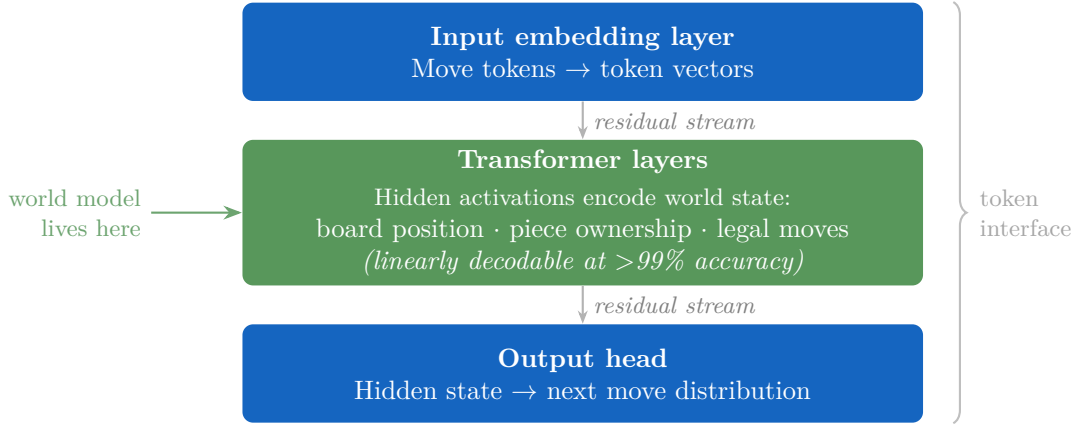

This framing predicts that LLMs should develop internal world-model representations in their hidden activations, and they do.
OthelloGPT \citep{li2023othello,nanda2023othello} is the cleanest demonstration:
trained only on move tokens, its hidden activations encode the full board state linearly at $>$99\% accuracy.
The world model is in the \emph{activations}, not in the tokens;
the tokens are the interface ($A = V$), while the latent world state lives in the hidden states (Figure~\ref{fig:othello}).

The same structure appears at scale.
\citet{gurnee2024space} show Llama-2 encodes linear representations of geographic space and calendar time.
\citet{dong2025planning} show prompt-level hidden states encode global attributes of the entire future response, not just the next token.
The
consistent pattern is that LLMs develop internal world representations far richer than their token-level objective requires.

\section{The Continuous Spectrum}

\begin{quote}
\textbf{Claim:} There is a natural continuous spectrum running from NTP to JEPA, with each step relaxing exactly one LLM constraint.
Moving along this spectrum also progressively sacrifices the two practical advantages that make LLMs trainable at scale: internet-scale self-supervised
data, and a well-matched transformer architecture.
\end{quote}

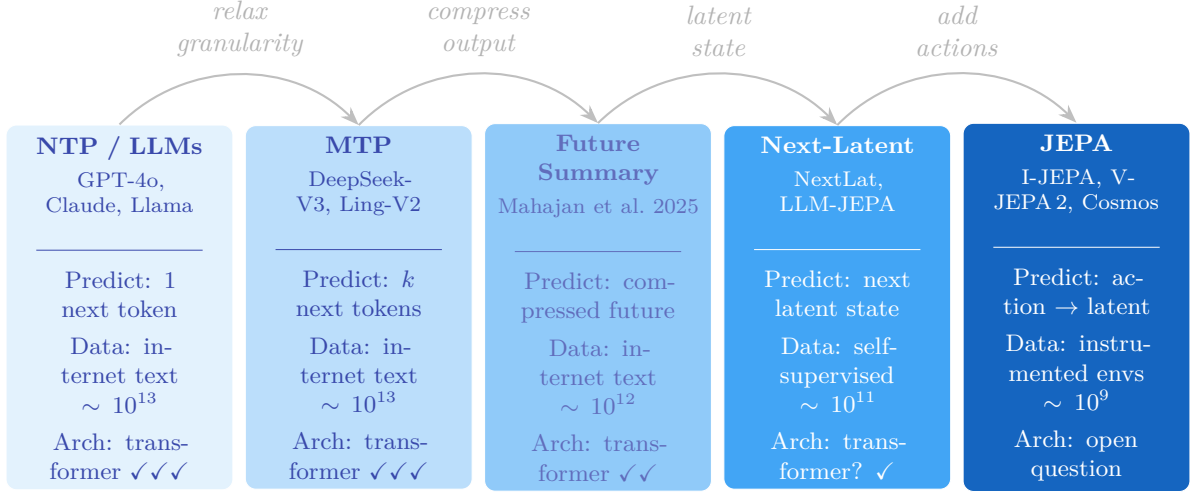
\begin{figure}[h]
\centering
\begin{tikzpicture}[
  node distance=0.15cm,
  spec/.style={rectangle, rounded corners=4pt, minimum width=3.0cm,
               minimum height=3.8cm, align=center, font=\footnotesize,
               text width=2.65cm, draw=white!60, inner sep=4pt},
  arr/.style={-{Stealth[length=7pt]}, thick, gray!50},
  arrlbl/.style={font=\small\itshape, text=gray!60, align=center}
]
  \node[spec, fill=specA, text=llmblue]    (ntp)  {
    \textbf{NTP / LLMs}\\[3pt]
    \scriptsize GPT-4o, Claude, Llama\\[5pt]
    \rule{2.2cm}{0.3pt}\\[3pt]
    \footnotesize Predict: 1 next token\\[3pt]
    Data: internet text\\$\sim\!10^{13}$\\[3pt]
    Arch: transformer \checkmark\checkmark\checkmark
  };
  \node[spec, fill=specB, text=llmblue,    right=of ntp] (mtp)  {
    \textbf{MTP}\\[3pt]
    \scriptsize DeepSeek-V3, Ling-V2\\[5pt]
    \rule{2.2cm}{0.3pt}\\[3pt]
    \footnotesize Predict: $k$ next tokens\\[3pt]
    Data: internet text\\$\sim\!10^{13}$\\[3pt]
    Arch: transformer \checkmark\checkmark\checkmark
  };
  \node[spec, fill=specC, text=llmblue!80, right=of mtp] (fs)   {
    \textbf{Future Summary}\\[3pt]
    \scriptsize Mahajan et al.\ 2025\\[5pt]
    \rule{2.2cm}{0.3pt}\\[3pt]
    \footnotesize Predict: compressed future\\[3pt]
    Data: internet text\\$\sim\!10^{12}$\\[3pt]
    Arch: transformer \checkmark\checkmark
  };
  \node[spec, fill=specD, text=white,      right=of fs]  (nl)   {
    \textbf{Next-Latent}\\[3pt]
    \scriptsize NextLat, LLM-JEPA\\[5pt]
    \rule{2.2cm}{0.3pt}\\[3pt]
    \footnotesize Predict: next latent state\\[3pt]
    Data: self-supervised\\$\sim\!10^{11}$\\[3pt]
    Arch: transformer? \checkmark
  };
  \node[spec, fill=specE, text=white,      right=of nl]  (jepa) {
    \textbf{JEPA}\\[3pt]
    \scriptsize I-JEPA, V-JEPA\,2, Cosmos\\[5pt]
    \rule{2.2cm}{0.3pt}\\[3pt]
    \footnotesize Predict: action $\to$ latent\\[3pt]
    Data: instrumented envs\\$\sim\!10^{9}$\\[3pt]
    Arch: open question
  };

  \draw[arr] (ntp.north)  to[bend left=45]
    node[above=3pt, pos=0.5, arrlbl]{relax\\granularity} (mtp.north);
  \draw[arr] (mtp.north)  to[bend left=45]
    node[above=3pt, pos=0.5, arrlbl]{compress\\output} (fs.north);
  \draw[arr] (fs.north)   to[bend left=45]
    node[above=3pt, pos=0.5, arrlbl]{latent\\state} (nl.north);
  \draw[arr] (nl.north)   to[bend left=45]
    node[above=3pt, pos=0.5, arrlbl]{add\\actions} (jepa.north);
\end{tikzpicture}
\caption{The spectrum from LLMs to JEPA.
Each node shows the prediction objective, training data scale, and architecture fit.
Moving right relaxes one world-model constraint but also degrades both LLM practical advantages.}
\label{fig:spectrum}
\end{figure}

\subsection{Step-by-Step Analysis}

Figure~\ref{fig:spectrum} shows the five stations of the spectrum; the following paragraphs examine each transition.

\paragraph{NTP $\to$ MTP.} This step relaxes the ``one token per step'' constraint.
\citet{gloeckle2024mtp}
show that predicting the next $k$ tokens simultaneously via $k$ independent heads improves reasoning and code performance (+15\% on MBPP; adopted in DeepSeek-V3).
\citet{zhong2026mtp}
provide the theoretical mechanism: MTP promotes convergence toward internal belief states via representational contractivity.
This step costs nothing in data or architecture: same internet-scale text, same transformer, only $k-1$ extra output heads.
It is a near-free upgrade.

\paragraph{MTP $\to$ Future Summary.}
This step decouples the prediction target from the token space.
\citet{mahajan2025future}
train an auxiliary head to predict a compressed representation of the long-term future (bag-of-words or reverse-LM embedding), improving maths and reasoning at 3B--8B scale.
Training data remains internet text with modest preprocessing; architecture is unchanged.

\paragraph{Future Summary $\to$ Next-Latent.}
This step moves the prediction target fully out of token space.
\citet{teoh2025nextlat} propose NextLat:
a transformer trained to predict its own next latent state, improving planning performance and inference speed (up to 3.3$\times$).
Crucially, this step can still train on internet-scale text; supervision comes from the model's own hidden states.
However, the architecture fit weakens.
Predicting continuous latent vectors rather than discrete tokens requires diffusion-style output heads or other adaptations,
and the transformer's inductive bias is no longer perfectly matched.

\paragraph{Next-Latent $\to$ JEPA.}
This is where both practical advantages collapse simultaneously.
\citeauthor{lecun2022path}'s JEPA predicts the latent state of a future observation conditioned on an external agent action.
Training now requires paired (observation, action, next observation) sequences from instrumented environments, orders of magnitude scarcer than text ($\sim$10$^9$ samples vs $\sim$10$^{13}$ tokens).
The right architecture is also an open question.
Existing JEPA models (I-JEPA, V-JEPA~2) work around this by re-discretising inputs as image or video patches,
effectively moving back toward the discrete-token end of the spectrum.
Whether the transformer generalises to truly continuous action-conditioned dynamics remains unresolved.

\subsection{The Data Question}

The first three steps on the spectrum (NTP, MTP, and Future Summary) all train on internet text, either directly or with modest preprocessing.
Even Next-Latent prediction can use internet-scale corpora:
the supervision signal comes from the model's own hidden states, not from external labels.
This is a crucial observation: moving to latent-state prediction does not require giving up internet-scale data, only changing the prediction target.

The real data cliff is at the final step.
JEPA requires paired (observation, action, next observation) sequences from instrumented environments:
robotics rigs, driving simulators, game engines, or video with inferred agent actions.
Such data is orders of magnitude scarcer than text ($\sim$10$^9$ samples vs $\sim$10$^{13}$ tokens).
V-JEPA~2 \citep{bardes2025vjepa2} approximates this by treating each video frame transition as an implicit action.
This is a workaround: true action-conditioned world models require environments that expose which action caused each transition.

The practical bottleneck therefore \emph{shifts} as you move up the spectrum:
at the LLM end it is compute; at the JEPA end it is data collection and environment instrumentation.
This asymmetry is underappreciated in debates that frame the gap as purely architectural.

\subsection{The Architecture Question}

The transformer's success with text is not accidental;
the architecture and the task were co-designed.
Self-attention over discrete positional embeddings, parallelised teacher-forcing, and a softmax prediction head are all engineered for sequences of discrete tokens.

So far, the transformer has extended further than expected, but via a recurring workaround: re-discretisation.
Vision Transformers \citep{dosovitskiy2021vit} split images into fixed-size patches and treat each as a token.
V-JEPA~2 does the same for video.
In both cases the transformer processes a \emph{discretised approximation} of a continuous input, not a truly continuous state.
The architecture's strength remains coupled to the discrete-token assumption.

At the far end of the spectrum, several alternatives have been proposed:
state-space models (Mamba) for explicit continuous-state recurrence;
diffusion transformers (DiT) for continuous output prediction; hierarchical architectures for multi-scale temporal planning.
The hypothesis advanced here is that what is missing is an analogous \emph{moment of crystallisation}:
a single architecture as well-matched to continuous sequential prediction as the transformer is to discrete tokens.
The field may be at a similar stage to NLP before 2017:
multiple competing approaches, each solving part of the problem, waiting for the unifying abstraction.

The data and architecture gaps interact:
a better architecture for continuous-state prediction could reduce the need for action-labelled data by learning more efficiently from self-supervised signals.
Progress likely requires advances on both fronts simultaneously.

\subsection{LeCun's Critique Revisited}

\citeauthor{lecun2022path}'s (\citeyear{lecun2022path}) argument has two parts:
(1) LLMs predict tokens, not world states, so they cannot plan;
(2) JEPA is the right alternative paradigm.
The spectrum view dissolves both.

\paragraph{On (1).}
The ``tokens, not states'' claim conflates the \emph{interface} with the \emph{representation}.
As Section~\ref{sec:empirical} shows, OthelloGPT and chess models encode rich world-state structure linearly in the hidden activations, not in the tokens.
The tokens are the action space $A = V$; the world model is in the hidden states.
LeCun's critique applies to the interface, not the internal representation.

\begin{figure}[h]
\centering
\begin{tikzpicture}
\begin{axis}[
  width=12cm, height=8cm,
  xlabel={State type},
  ylabel={Planning horizon},
  xmin=0, xmax=1, ymin=0, ymax=1,
  xtick={0,0.5,1}, ytick={0,0.5,1},
  xticklabels={Discrete, , Continuous},
  yticklabels={Single-step, , Multi-step},
  grid=both, grid style={line width=0.3pt, draw=gray!30},
  axis line style={gray!50},
  tick style={gray!50},
  label style={font=\small},
  tick label style={font=\small},
  clip=false
]
  \node[font=\footnotesize\itshape, text=gray!60, align=center] at (0.25,0.25) {Standard NTP};
  \node[font=\footnotesize\itshape, text=gray!60, align=center] at (0.25,0.75) {Long-horizon\\token planners};
  \node[font=\footnotesize\itshape, text=gray!60, align=center] at (0.75,0.75) {Full world models};
  \node[font=\footnotesize\itshape, text=gray!60, align=center] at (0.75,0.25) {Latent one-step};

  \addplot[only marks, mark=*, mark size=3pt, color=llmblue]
    coordinates {
      (0.10, 0.10)
      (0.15, 0.40)
      (0.18, 0.58)
      (0.42, 0.64)
      (0.68, 0.62)
      (0.86, 0.82)
      (0.90, 0.90)
    };

  \node[font=\scriptsize, anchor=south west] at (0.10,0.10) {\;Standard LLMs};
  \node[font=\scriptsize, anchor=south west] at (0.15,0.40) {\;Chain of Thought};
  \node[font=\scriptsize, anchor=south west] at (0.18,0.58) {\;MTP};
  \node[font=\scriptsize, anchor=south west] at (0.42,0.64) {\;Future Summary};
  \node[font=\scriptsize, anchor=south west] at (0.68,0.62) {\;Next-Latent};
  \node[font=\scriptsize, anchor=south west] at (0.86,0.82) {\;LLM-JEPA};
  \node[font=\scriptsize, anchor=south west] at (0.90,0.90) {\;JEPA / V-JEPA\,2};

\end{axis}
\end{tikzpicture}
\caption{Architectures in the (state type $\times$ planning horizon) design space.
The dashed diagonal traces the spectrum of Section~3.
CoT sits directly above Standard LLMs;
it increases planning horizon without changing the state type.
The spectrum reveals a densely populated intermediate region that LeCun's binary framing overlooks.}
\label{fig:quadrant}
\end{figure}
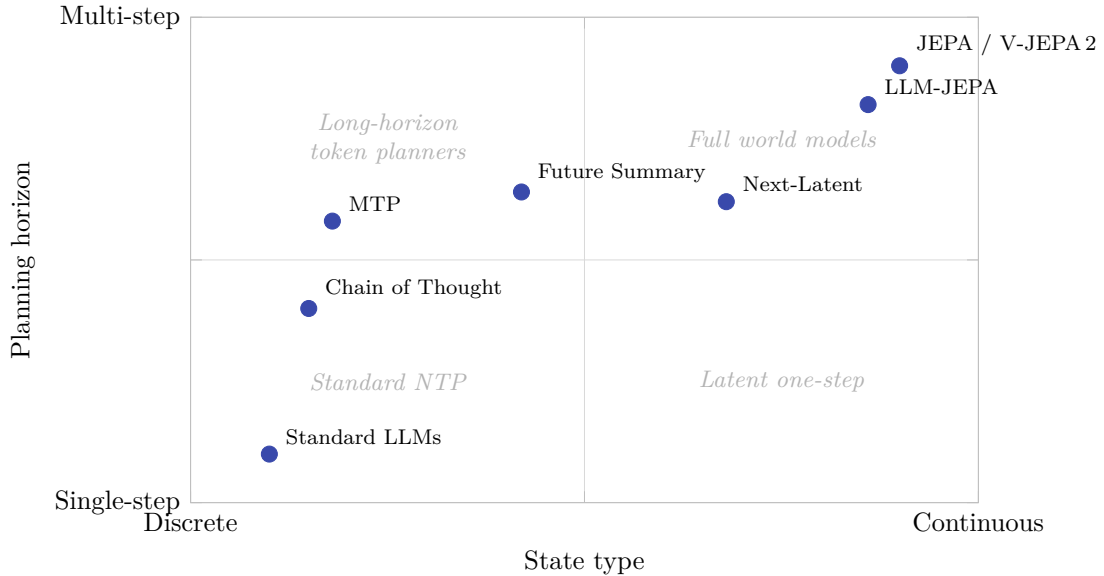

\paragraph{On (2).}
Chain-of-thought (CoT) and extended thinking (DeepSeek-R1, OpenAI o3) can be understood as LLMs moving along the spectrum's step-granularity axis \emph{without architectural change}.
Generating intermediate reasoning tokens increases the effective planning horizon within the token-state world model.
\citet{dong2025planning} show that even without CoT, the hidden state already encodes a rough plan for the full response.
JEPA is not a different paradigm; it is the far end of the same spectrum (Figure~\ref{fig:quadrant}).

\section{Discussion and Conclusion}

The preceding sections have established that LLMs are world models (constrained ones) and that world models are their generalisation.
The practical consequence is that the question is not \emph{whether} to abandon LLMs for world models, but \emph{how far along the spectrum} a given task requires.
Multi-step reasoning tasks may be well-served by MTP or CoT;
long-horizon physical simulation requires the JEPA end.
Each step along the spectrum is also a natural initialisation for the next, suggesting an incremental transition rather than a cold start.

The two open questions that define the research frontier are treated in Sections~3.2 and~3.3.
The intermediate spectrum steps (MTP through Next-Latent) are particularly attractive because they preserve both LLM practical advantages.
Specifically, internet-scale self-supervised training and a well-matched transformer architecture.
Meanwhile, these intermediate steps improve world-model capacity.
The true break comes at the JEPA step, where both advantages disappear simultaneously:
training data becomes scarce and the right architecture is unknown.
What may be needed is not just more data but a new architectural primitive for continuous-state prediction, analogous to what the transformer was for text.

LeCun is right that the ultimate goal (action-conditioned latent-space world models capable of closed-loop planning) is beyond current LLMs.
But the path there is gradient ascent, not a discontinuous jump.
The open questions are what data and what architecture; the destination is already clear.

\bigskip
\noindent\textit{This opinion paper was written with the assistance of a constrained to natural language special case world model.}

\bibliographystyle{plainnat}
\bibliography{references}

\end{document}